%% file: main.tex
\def\year{2022}\relax
\documentclass[letterpaper]{article} 
\usepackage{aaai22}  
\usepackage{times}  
\usepackage{helvet}  
\usepackage{courier}  
\usepackage[hyphens]{url}  
\usepackage{graphicx} 
\usepackage{natbib}  
\usepackage{caption} 
\DeclareCaptionStyle{ruled}{labelfont=normalfont,labelsep=colon,strut=off} 
\usepackage{algorithm}
\usepackage{algorithmic}

\nocopyright
\usepackage{newfloat}
\usepackage{listings}
\floatstyle{ruled}
\newfloat{listing}{tb}{lst}{}
\floatname{listing}{Listing}

\usepackage{amsfonts}

\usepackage{xcolor}
\usepackage{amsmath}
\usepackage{array}
\usepackage{booktabs}
\usepackage{multirow}
\usepackage{bm}
\usepackage{amssymb}
\usepackage{pifont}

\definecolor{darkergreen}{RGB}{19, 182, 53}
\definecolor{red2}{RGB}{252, 54, 65}
\newcommand\redp[1]{\textcolor{red2}{(#1)}}
\newcommand\greenp[1]{\textcolor{darkergreen}{(#1)}}
\newcolumntype{P}[1]{>{\centering\arraybackslash}p{#1}}

\newlength\savewidth\newcommand\shline{\noalign{\global\savewidth\arrayrulewidth
  \global\arrayrulewidth 1pt}\hline\noalign{\global\arrayrulewidth\savewidth}}

\title{Suppressing Static Visual Cues via Normalizing Flows for Self-Supervised Video Representation Learning}
\author{
    Manlin Zhang\textsuperscript{\rm 1}\equalcontrib,
    Jinpeng Wang\textsuperscript{\rm 2}\equalcontrib,
    Andy J. Ma\textsuperscript{\rm 1,\rm 3}\equalcontrib\thanks{Corresponding author.}
}
\affiliations{
    \textsuperscript{\rm 1}School of Computer Science and Engineering, Sun Yat-sen University, Guangzhou, China\\
    \textsuperscript{\rm 2}School of Electronics and Information Technology, Sun Yat-sen University, Guangzhou, China\\
    \textsuperscript{\rm 3}Key Laboratory of Machine Intelligence and Advanced Computing, Ministry of Education, China\\
    \{zhangmlin3, wangjp23\}@mail2.sysu.edu.cn, majh8@mail.sysu.edu.cn
}

\begin{document}

\maketitle

\input{sections/0-Abstract}
\input{sections/1-Introduction}
\input{sections/2-RelatedWork}
\input{sections/3-ProposedMethod}
\input{sections/4-Experiments}
\input{sections/5-Conclusion}

\section{Acknowledgement}
This work was supported partially by National Natural Science Foundation of China (No. 61906218), Guangdong Basic and Applied Basic Research Foundation (No. 2020A1515011497) and Science and Technology Program of Guangzhou (No. 202002030371).

\bibliography{main.bib}

\end{document}


\maketitle

\section{Pseudo Code}
The pseudo-code of the proposed method to generate motion-preserved videos for suppressing static visual cues is presented in Algorithm~\ref{algo:bg_suppress_detail} as follows.
\input{sections/S_3_pseudo_code}

\section{Implementation Details}
\noindent\textbf{Network.}
The backbones used for experiments are the S3D and 3D Resnet-18 (R3D).
During pre-training, we attach a projection head to the last convolutional (conv) layer, i.e. the \textit{block5} for the S3D and the \textit{res4} for 3D Resnet-18.
The projection head consists of one global spatio-temporal average pooling and one fully connected layer with output dimension 128.
During evaluation, the projection head is discarded and replaced by a linear classifier.

\noindent\textbf{Pipeline.}
Before suppressing static cues, the spatial resolution of each video frame is down-sampled to $64\times 64$ for computational efficiency.
Then, the down-sampled frames are fed into the flow-based generative model pre-trained on ImageNet as in~\cite{advflow}) to obtain latent variables, for suppressing static cues.

\noindent\textbf{Momentum Dictionary.}
The size of the momentum dictionary for pre-training on UCF101 is set to 2048.
As Kinetics contains 200K videos, we use a larger memory bank with the size of 16,384 to save features.

\noindent\textbf{Basic Data Augmentation.}
The hyperparameters for the basic data augmentations are: color jittering (0.4, 0.4, 0.4, 0.1, p=0.8), gray scale (p=0.2), gaussian blur (p=0.5), horizontal flip (p=0.5). 
We conduct consistent augmentation for each frame in a video clip.

\section{Additional Experiments}
\subsection{Integrating with Other SSL Approach}
Besides the MoCo, the proposed S$^2$VC can be considered as a data augmentation technique and integrated with other self-supervised learning (SSL) methods, e.g., DPC~\cite{DPC}.
In this experiment, we follow the default settings of the DPC except for adding the proposed S$^2$VC for data augmentation.
We pre-train two models w/o and w/ suppressing static visual cues on the UCF101 dataset for 300 epochs.
As shown in Table~\ref{tab:dpc}, our method surpasses the DPC on both video retrieval and action recognition tasks, which demonstrates the effectiveness of our method.
\input{tables/S_5_DPC}

\subsection{Analysis on Bias Caused by Static Cues}
\noindent\textbf{Statistics of Static Information.}
In this experiment, we investigate the proportion of human/motion regions in natural videos.
The YOLOv5 method~(Jocher et al.~\citeyear{yolo_v5_doi}) is employed to detect human and the standard Farneback optical flow algorithm~\cite{farneback} is used to detect motion as illustrated in Fig.~\ref{fig:staticstic}(a).
By averaging the proportion of detected human regions frame-by-frame for videos, we obtained the mean human proportions on different datasets as shown in the first row of Fig.~\ref{fig:staticstic}(b). 
Likewise, the mean motion proportion is calculated and shown in the second row.
It can be observed that videos in all the three publicly available activity datasets have less than 20-percent of regions containing human, and have only about 30-percent of regions related to motion.
Both Statistics demonstrate motion information is \textit{less dominated} than static visual cues in videos.
\input{figures/S_1_Statistics}

\noindent\textbf{Natural/Shuffled Video Classification.}
We extend the natural/shuffled video classification experiment by pre-training on a larger-scale dataset Kinetics-400.
We use the 3D ResNet-18 as the backbone and compare our self-supervised pre-trained model with the open-source supervised pre-training method~\cite{r3d}.
As shown in Table \ref{tab:bin_test}, the proposed S$^2$VC method outperforms the supervised pre-trained model with a large margin for both small and large scale datasets.
This result indicates that the feature learned without dealing with the representation bias problem \textit{may still be wrongly guided by static visual cues} even when pre-training on a larger-scale dataset.
\input{tables/S_2_Binary_Classification}

\input{figures/S_6_Cos_Sim}

\input{figures/S_7_Visualize_Sample}
\subsection{Analysis on Suppressing Effects}
\noindent\textbf{The Cosine Similarity.}
Besides the \textit{Intra-class} comparison in the manuscript, the \textit{Intra-video} and \textit{Inter-class} comparison results are presented in this section.
Denote $V_A$ and $V_B$ as two videos from different action categories.
\textit{Intra-video}:
Computing the similarity frame-wise between adjacent frames in the same video $V_A$.
\textit{Inter-class}:
Computing the similarity between one frame from $V_A$ and the other frame from $V_B$.
Notice that only the spatial (static) visual similarity from frame vs. frame is evaluated, i.e., no temporal information is included.

The results are shown in Fig.~\ref{fig:intra_video_cos_sim} and Fig.~\ref{fig:inter_class_cos_sim}.
All these results demonstrate that the proposed S$^2$VC reduces the spatial visual similarity significantly.
Since static visual cues contribute to higher spatial similarity, these results validate that static cues are mostly erased in the motion-preserved latent vector $Z_p$.
Under the framework of contrastive learning, the learned features rely more on motion cues to better separate each video from others.

\noindent\textbf{Visualization Results.}
More visualization examples of the generated motion-preserved video are shown in Fig.~\ref{fig:aug_visualize}.
We consider different types of actions and show which parts of the frame are preserved after S$^2$VC.
We observe most spatial cues are weakened clearly, like the sports venue in the first row.
The preserved regions are highly correlated to human motion and are able to recognize the action categories according to the visual relation along the time dimension.

\input{figures/S_8_Norm_fit}
\subsection{Distribution Visualization}
In this section, we evaluate whether the pixel and latent space follows normal distribution, respectively.
For this purpose, we randomly sampled 8 pixels from each frame or latent variables conditioning on static factors in a video.
Then, we fit the normal distribution to each of the pixels or latent variables.
The fitting results, mean square error (MSE) and Kolmogorov-Smirnov test decision are shown in Fig.~\ref{fig:norm_fit}.
The MSE is to measure the normalized difference between the data and ground-truth normal distribution. 
For the Kolmogorov-Smirnov test, the data for each of the pixels or latent variables is first normalized by subtracting the mean and standard deviation.
Then, the normalized data is tested with the null hypothesis ($H_0$) that the data follows a standard normal distribution with significance level $\alpha = 0.05$.
From Fig.~\ref{fig:norm_fit}, we can see that the MSE of fitting normal distribution in pixel level is remarkably larger than that in the latent space.
For the results of the statistical test, we cannot reject the null hypothesis that latent variables conditioning on a video follow normal distribution in most cases.
Recall that the distribution of each latent variable encoded by normalizing flows should be one-dimensional (univariate) standard normal, if it is completely random without any other information given in advance.
However, when constrained in a video, the distribution is affected by the shared static cues, so that the standard normal distribution is shifted and scaled.
The fitting results in Fig.~\ref{fig:norm_fit} aligns with our analysis.

\bibliography{supplementary.bib}

%% file: sections/0-Abstract.tex
\begin{abstract}
Despite the great progress in video understanding made by deep convolutional neural networks, feature representation learned by existing methods may be biased to static visual cues.
To address this issue, we propose a novel method to suppress static visual cues (S$^2$VC) based on probabilistic analysis for self-supervised video representation learning.
In our method, video frames are first encoded to obtain latent variables under standard normal distribution via normalizing flows.
By modelling static factors in a video as a random variable, the conditional distribution of each latent variable becomes shifted and scaled normal.
Then, the less-varying latent variables along time are selected as static cues and suppressed to generate motion-preserved videos.
Finally, positive pairs are constructed by motion-preserved videos for contrastive learning to alleviate the problem of representation bias to static cues.
The less-biased video representation can be better generalized to various downstream tasks.
Extensive experiments on publicly available benchmarks demonstrate that the proposed method outperforms the state of the art when only single RGB modality is used for pre-training.
The code is available at \textcolor{blue}{\url{https://github.com/mettyz/SSVC}}.
\end{abstract}

%% file: sections/1-Introduction.tex
\section{Introduction}
Recent top-performing approaches to solving video understanding tasks are based on supervised learning with a large amount of labeled data for training. 
Due to the strong data fitting capacity of deep convolutional neural networks, competitive performance can be achieved for recognizing actions in videos~(Carreira et al.~\citeyear{I3D};~\citealt{zhang2021morphmlp}).
One of the key factors for the success may owe to the strong correlation between action class and object/background known as representation bias in~(Li et al.~\citeyear{li2018resound};~\citealt{choi2019can}).
For example, the action \textit{Riding Bike} could be recognized by the presence of the object \textit{Bike} and the action \textit{Swimming} is recognized by the scene \textit{water}.
Such representation bias in action datasets may provide shortcuts to solve the data-label fitting problem.
Nevertheless, the learned feature representation without proper motion modelling may be biased to static visual cues, which limits the generalization ability to recognize or detect actions requiring temporal reasoning.
\input{figures/1_Teaser}

To verify this issue, we first pre-trained the R3D network~(Hara et al.~\citeyear{r3d}) for feature extraction in two different ways.
The first one is supervised learning by using manual annotations on the UCF101 dataset~(Soomro et al.~\citeyear{ucf101}), while the second one is our self-supervised method trained on the same dataset by mitigating the representation bias.
Then, the learned feature representations are evaluated on the HMDB51 dataset~\cite{hmdb51}.
The downstream task is defined as simple temporal-order (natural/shuffled) classification as illustrated in Fig.~\ref{fig:teaser} to assess the generalization ability of the learned features.
Fig.~\ref{fig:teaser}(a) shows that motion information is suppressed while static visual cues are maintained by video shuffling.
Without dealing with the problem of representation bias, the supervised method performs worse than ours (\textbf{73.5\% v.s. 79.1\%} \greenp{5.6$\uparrow$}) in the downstream task of temporal-order classification (different from the recognition task used for pre-training).
This verifies our hypothesis that the generalization ability may degrade due to the misleading guidance by static visual cues.

In this paper, we propose a novel method to suppress static visual cues (S$^2$VC) for self-supervised video representation learning, such that the representation bias is mitigated.
Since the pixel space of each frame in a video is highly complicated with high dimensionality, it is not robust to directly extract static cues from it.
To estimate the distribution of the pixel space, each video frame is encoded to obtain a latent vector under multivariate standard normal distribution by using normalizing flows (NF).
However, when constrained to a specific video, each latent variable cannot be simply considered as one-dimensional  (univariate) standard normal.
We model static factors in a video as a random variable such that the conditional distribution of each latent variable becomes standard normal with shifting and scaling.
The standard deviation of the conditional distribution that reflects the correlation between latent variables and static factors is then empirically estimated to select static cues.
Based on probabilistic analysis, static cues are suppressed to generate motion-preserved videos by the invertibility of the NF model.
Such generated videos are treated as pseudo positives for contrastive learning to mitigate the representation bias w.r.t. static visual cues.

The contributions of this work are three-fold: 
\emph{i}. 
We develop a novel method to suppress static visual cues (S$^2$VC) via normalizing flows for self-supervised video representation learning, 
such that the problem of representation bias is mitigated with improved generalization ability.
\emph{ii}.
Based on probabilistic analysis, static cues are recognized and suppressed to generate motion-preserved videos for self-supervised pre-training.
\emph{iii}. 
Extensive experiments with quantitative and qualitative evaluation demonstrate the effectiveness of our method on various downstream tasks.

%% file: figures/1_Teaser.tex
\begin{figure}[ht]
	\centering
	\includegraphics[width=.9\linewidth]{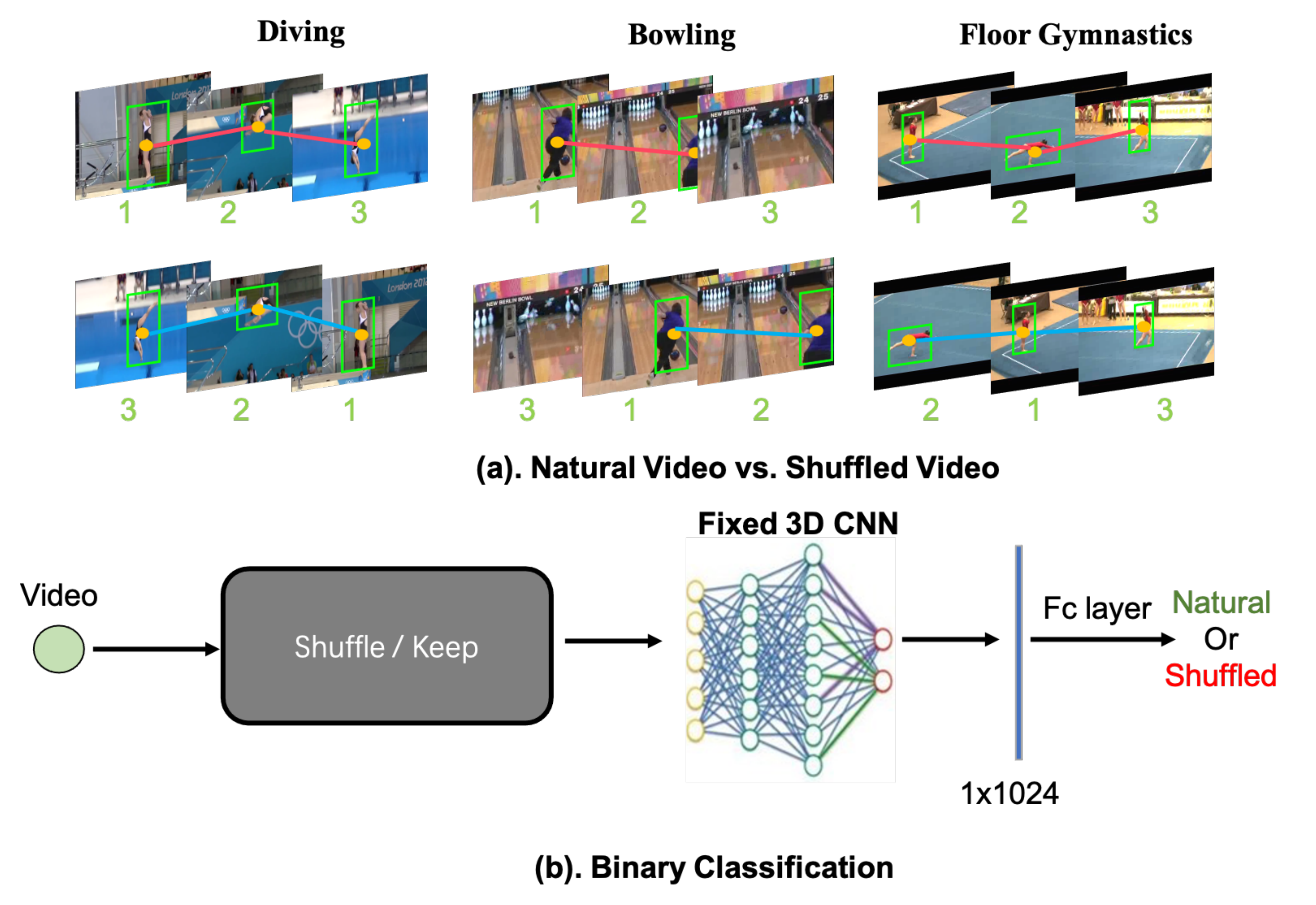}
	\caption{\textbf{Verifying the importance of motion cues}.
	(a): Natural videos (top row) and shuffled videos (bottom row).
	(b): With a fixed-weight 3D CNN, a linear classifier is trained to distinguish natural videos from the shuffled ones.
	We pre-train the R3D as feature extractor on the UCF101 by two different methods, i.e., supervised learning and our self-supervised learning method w/o and w/ suppressing static visual cues, respectively.
	The performance of classifying natural/shuffled videos on HMDB51 is \textbf{73.5\%} by using the former and \textbf{79.1\%} \greenp{5.6$\uparrow$} by using the latter.
	Notice that both the test dataset and downstream task are different from those used for training, which implies better generalization ability of the learned representations by suppressing static cues.
	}
 	\label{fig:teaser}
\end{figure}

%% file: sections/2-RelatedWork.tex
\section{Related Work}
\noindent\textbf{Self-supervised Video Representation Learning} aims at learning visual representations without using manually-annotated labels.
Existing methods for video representation learning can be divided into two categories.
The first one is to design pretext tasks, in which pseudo labels are automatically generated from videos for training.
Representative methods along this line include predicting rotation~\cite{jing2018self}, cloze~\cite{st_cloze}, clip order~(Misra et al.~\citeyear{shuffle_and_learn};~\citealt{lee2017unsupervised, VCOP}), playback speed~(\citealt{speednet}; Wang et al.~\citeyear{pace};~\citealt{PRP,RSPNet}) and so on.
The second category is based on contrastive learning which has recently achieved great success in the image domain~\cite{moco,moco_v2,SimCLR}.
The key idea is to train a feature extractor that makes a training sample similar to its generated positives and dissimilar to its negatives in the embedding space.
Existing methods have been proposed to generate positive pairs by video clips sampled from the same video~\cite{CVRL,wang2021enhancing,lin2021learning}, or codes from the same position of adjacent frames~(Han et al.~\citeyear{DPC},~\citeyear{MemDPC}).
Since additional modalities are available in videos, positive pairs can also be determined by audio~(\citealt{owens2018audio,XDC};~Korbar et al.~\citeyear{korbar2018cooperative}), text~\cite{sun2019videobert}, or optical flow~(Han et al.~\citeyear{han2020self}).
Though existing methods show improved performance for downstream tasks, they may be still biased to static visual cues like background or non-moving objects.
To solve this problem, this paper proposes to generate motion-preserved videos by normalizing flows for less-biased representation learning.

\input{figures/3_1_PPL}

\noindent\textbf{Flow-based Generative Model} is one of the widely used approaches for data generation developed with strong theory in probability~(\citealt{INN}; Dinh et al.~\citeyear{nice}; Dinh et al.~\citeyear{realnvp}).
It builds on a series of invertible and differentiable functions that transforms the highly-complicated raw data distribution to the simple and interpretable standard normal distribution.
This transforming sequence is called normalizing flows (NF) and is served as the foundation of invertible neural network.
In recent years, NF has been successfully deployed in many applications including image generation~\cite{glow}, compression~\cite{xiao2020invertible},  colorization~\cite{cINN}, adversarial attack~(Dolatabadi et al.~\citeyear{advflow}), minimally invasive surgery~\cite{medical_app}, etc.
To the best of our knowledge, this work is the first to suppress static cues in videos by using NF for self-supervised learning.

%% file: figures/3_1_PPL.tex
\begin{figure*}[t]
	\centering
	\includegraphics[width=0.96\linewidth]{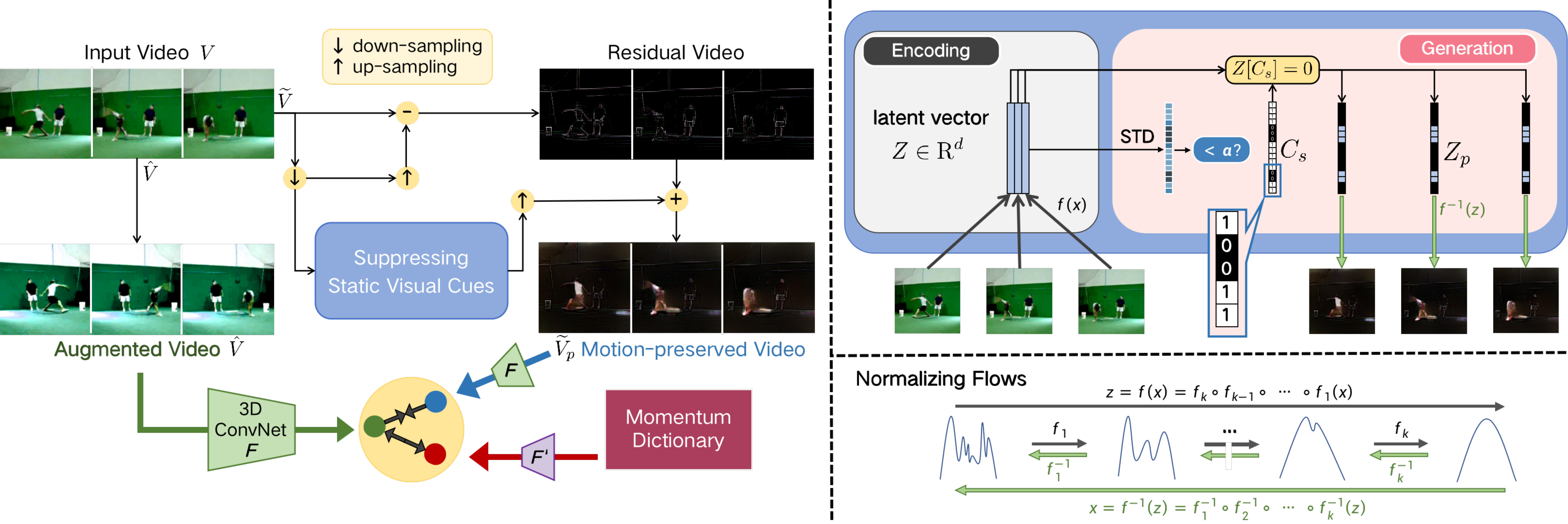}
	\caption{
	Left: \textbf{Training pipeline}.
	We propose to learn video representations by training a 3D CNN to encode similar features for the motion-preserved and the augmented video.
	The motion-preserved video retains the \textit{residual} and \textit{motion} information, while clearly wiping the static cues.
	Right: \textbf{Suppressing static visual cues}.
	We illustrate the suppressing process with two sub-process as shown on the right.
	In the encoding process, each video frame is mapping to a latent vector $Z$ via normalizing flows. 
	In the generation process, static cues are first selected by thresholding the standard deviation of the latent vectors w.r.t the temporal dimension.
	Then, the selected channels are set to zero to obtain $Z_p$ for generating motion-preserved video.
	}
 	\label{fig:framework}
\end{figure*}

%% file: sections/3-ProposedMethod.tex
\section{Methodology}
The objective of our proposed method is to mitigate the representation bias brought by the strong correlation between actions and static visual cues, such that the learned features can be better generalized to different kinds of downstream tasks.
The rationale is to perform probability-based video transformations that preserve motion information but suppress static visual cues (S$^2$VC) for videos.
In the following subsections, we first introduce the overall architecture of the proposed method.
Then, details are given to elaborate the idea of the proposed S$^2$VC for motion-preserved video generation via normalizing flows~(NF).
At last, we present the way to integrate the novel S$^2$VC with existing self-supervised methods for video representation learning.

\subsection{Overall Architecture}\label{architect}

The training pipeline of our method is shown on the left of Fig.~\ref{fig:framework}. 
For a given unlabelled input video $V$, we start with two random augmentations and get $\hat{V}=\hat{s}(V)$ and $\widetilde{V}=\widetilde{s}(V)$ respectively, where $\hat{s}$ and $\widetilde{s}$ are randomly sampled from the basic data augmentation set $S$. 
It consists of (e.g.) random cropping, random horizontal flip, color jittering and Gaussian blur.
One of the randomly augmented videos $\widetilde{V}$ is used to generate the motion-preserved video $\widetilde{V_p}$ by suppressing static visual cues via normalizing flows.
To mitigate the computational cost, a spatially down/up-sampling process is performed before/after the flow-based generative model.
The information loss is compensated by the residual video.
After that, $\hat{V}$ and $\widetilde{V}_p$ are fed into the 3D backbone $F$ for feature extraction to obtain $v = F(\hat{V})$ and $v_p = F(\widetilde{V}_p)$.
The feature extractor $F$ is learned by minimizing the distance between $v$ and $v_p$, and maximizing the distance between $v$ and other pseudo negatives from the momentum dictionary, whose features are extracted by the momentum encoder $F'$.

\subsection{Suppressing Static Visual Cues}
\label{sec:adv_attack}
The proposed method for suppressing static visual cues is illustrated on the right of Fig.~\ref{fig:framework}.
For an input video, each frame is first encoded to obtain latent variables under standard normal distribution by normalizing flows (NF).
Then, the motion-preserved video is generated by suppressing the less-varying latent variables (static cues) along time.
Details on these two steps are provided as follows.

\noindent
\textbf{Encoding Video Frames via Normalizing Flows.}
Denote the vectorized frames in the input video as $X_1, \cdots, X_L \in \mathbb{R}^{d}$, where $L$ is the number frames and $d$ is the product of image height, width and channels.
These $d$-dimensional vectors can be considered as a sequence of observations for a random vector $X$ with the probability density $p_X$.
Since the dimension of the random vector $X$ is very high, it is intractable to directly estimate the density $p_X$ correctly.
Moreover, $p_X$ is highly complicated due to variations like camera motions and illumination changes in videos.
Without accuracy estimation of the data distribution, it not robust to extract static cues directly from the raw observed data.
As a result, we propose to estimate the density $p_X$ of the high-dimensional random vector $X$ by normalizing flows (NF).

The idea of NF~(Kobyzev et al.~\citeyear{NF_review}) is depicted on the right of Fig.~\ref{fig:framework}.
A sequence of simple invertible transformations $f_1, \cdots, f_k$ (e.g., affine coupling and channel-wise permutation/convolution) maps $X$ to the latent random vector $Z$, which has the same dimension as $X$.
Denote the composition function of $f_1, \cdots, f_k$ as $f$, i.e.,  $f = f_k \circ \cdots \circ f_1$.
The mapping $f$ from $X \in \mathbb{R}^{d}$ to $Z \in \mathbb{R}^{d}$ is invertible and differentiable.
By using $f$, $X$ from the highly complicated distribution can be transformed to $Z$ in a straightforward predefined distribution such as multivariate standard normal.
To determine the parameters $\theta$ in the mapping function $f$, the density $p_X$ is rewritten by the change-of-variables rule as,
\begin{equation}
\begin{aligned}
    p_X(X) = p_Z\left( f_\theta(X) \right) \left| \det (D f_\theta) (X) \right|
\end{aligned}
\end{equation}
where $p_Z$ is the probability density of the latent random vector $Z$ and $\det(D f_\theta)(X)$ denotes the determinant of the Jacobian matrix of partial derivatives of $f_\theta$ over $X$.
Given an image dataset $\mathcal{D}_{\textrm{NF}}$ (e.g., ImageNet), the model parameters $\theta$ are learned by maximizing the log-likelihood as follows,
\begin{equation}
\begin{aligned}
    \max_\theta \mathbf{E}_{X \sim \mathcal{D}_{\textrm{NF}}} \left( \log p_Z\left( f_\theta(X) \right) + \log \left| \det (D f_\theta) (X) \right| \right)
\end{aligned}
\end{equation}
where $\mathbf{E}$ is the mathematical expectation.

In our method, the predefined density $p_Z$ is set to multivariate standard normal as in~(Dinh et al.~\citeyear{realnvp}),
i.e., $Z \sim \mathcal{N}(\mathbf{0}, I)$, where $\mathbf{0}$ is a $d$-dimensional zero vector and $I$ is a $d \times d$ unit matrix.
With the pre-trained flow model, the vectorized frames $X_1, \cdots, X_L$ are mapped into the latent space to obtain $Z_1, \cdots, Z_L$, which are used to detect the temporally-varying patterns and extract static cues.

\noindent
\textbf{Motion-preserved Video Generation.}
Since the latent vector $Z$ follows $d$-dimensional standard normal distribution $\mathcal{N}(\mathbf{0}, I)$, latent variables in $Z$ are independent with each other.
Thus, we propose to analyze each latent variable $Z^i, i \in \{ 1, \cdots, d \}$ in $Z$ to identify static cues separately. 
If $Z^i$ is completely random without any other information given in advance, it is obvious that the distribution of each latent variable $Z^i$ is one-dimensional (univariate) standard normal, i.e., $Z^i \sim \mathcal{N}(0, 1)$.
Nevertheless, when the latent variable $Z^i$ is constrained to be in a certain video, the completely random assumption is not valid.

We regard static factors (e.g., background, scene) inherited in the input video affecting the distribution of each $Z^i$ as a random variable $Y$.
For selection of static cues, the objective is to determine the density $p_{Z^i|Y}$ of $Z^i$ conditioning on $Y$.
Let the dependence between $Z^i$ and $Y$ be modelled by the correlation coefficient $\rho_i$.
To make the marginal density $p_{Z^i}$ standard normal, the joint density $p_{Z^i,Y}$ is assumed to be two-dimensional (bivariate) normal for maximum entropy.
Denote $(Z^i,Y) \sim \mathcal{N}(0, \mu, 1, \sigma^2, \rho_i)$, where $\mu, \sigma^2$ are the mean and variance of $Y$ respectively.
According to properties of normal conditional distribution~\cite[268-269]{probability_ross_textbook}, the conditional density $p_{Z^i|Y=y}$ for a given value of $Y = y$ is still normal and can be written as,
\begin{equation}\label{prob_cond}
\begin{aligned}
    (Z^i|Y=y) \sim \mathcal{N}(\frac{1}{\sigma} \rho_i (y - \mu), 1-\rho^2_i)
\end{aligned}
\end{equation}
This equation implies that the latent variable $Z^i$ conditioning on an input video can be considered as standard normal random variable with shifting and scaling.
With the condition $Y = y$, the mean and variance are changed to $ \rho_i (y - \mu) / \sigma$ and $1-\rho^2_i$, respectively.

For a latent variable $Z^i$ strongly correlated with static factors represented by $Y$, the dependence modelled by the correlation $\rho_i$ between $Z^i$ and $Y$ is large.
According to eq.~\eqref{prob_cond}, this means the variance $1-\rho^2_i$ is small for the latent variable $Z^i$ encoding static cues.
Notice that the variance $1-\rho^2_i$ is independent of the value of $Y=y$.
The variance or standard deviation (STD) can be estimated empirically by the observations $Z_{1}^i,\dots,Z_{L}^i$ of the latent variable $Z^i$ in a video.
Denote the STD of the conditional density $p_{Z^i|Y=y}$ as $\sigma_{Z^i|Y}$.
We propose to select the set $C_s$ of latent variables with small empirical STDs as static cues, i.e.,
\begin{equation}\label{select_sc}
\begin{aligned}
    C_s = \{i | \sigma_{Z^i|Y} \approx  \textrm{STD}(Z_{1}^i, \dots, Z_{L}^i) < \alpha \}
\end{aligned}
\end{equation}
where $\alpha$ is the threshold hyperparameter used to decide whether the $i$-th latent variable is selected or not.

Let the latent vector that preserves motion information but suppresses static cues be $Z_p$.
For $i \in C_s$, $Z_p^i$ is set to $\rho_i (y - \mu) / \sigma$ with the highest probability density, i.e., the mean of the conditional distribution $Z^i|Y=y$ as derived in eq.~\eqref{prob_cond}.
In this way, the variance of $Z_p^i$ is equal to 0 for minimum (zero) information entropy to suppress static cues. 
Since the marginal density $p_Y$ takes the maximum value at $Y = \mu$, we set $Z^i_p = 0$ for $i \in C_s$ by substituting $y = \mu$ into $\rho_i (y - \mu) / \sigma$.
For $i \notin C_s$, motion cues are preserved by setting $Z^i_p = Z^i$.
Due to invertibility of the NF model $f_\theta$, each frame in the motion-preserved video is generated by,
\begin{equation}
    X_p = f_\theta^{-1}(Z_p)
\end{equation}
The pseudo-code of our method is given in the supplementary material.
Other strategies to suppress static cues are also presented for comparison in the ablation study.

\noindent
\textbf{Discussion on Generative Models.}

\emph{i}. 
The generative adversarial network (GAN) has achieved success in the literature by jointly training a generator and a discriminator in an adversarial manner~\cite{GAN,Bi_cGAN}.
In most existing methods based on GAN, there is no encoder to transform the image modality into the latent space.
Though the generator in GAN could be used for encoding, the generation results are without explicit probability interpretation.
Hence, it is difficult if not impossible to suppress static cues by the GAN approach.

\emph{ii}. 
Different from GAN, the variational auto-encoder (VAE)~\cite{VAE,cVAE} can encode an input image $X$ to a latent vector $Z$ under a multivariate normal distribution $\mathcal{N}(\bm{m}_X, \textrm{diag}(\bm{\sigma}_X^2))$.
The mean vector $\bm{m}_X$ and standard deviation vector $\bm{\sigma}_X$ are determined by learnable parameters and the input image $X$.
The latent vector $Z$ is obtained by randomly sampling from $\mathcal{N}(\bm{m}_X, \textrm{diag}(\bm{\sigma}_X^2))$ and can be written as $Z = \bm{m}_X + \bm{\sigma}_X \odot \bm{\epsilon}$, where $\bm{\epsilon} \sim \mathcal{N}(\bm{0},I)$ and $\odot$ is the element-wise product.
Due to the randomness in computing the observations of $Z$, the encoded vectors $Z_1, \cdots,  Z_L$ in a video may not be able to preserve the continuity of the input frames over time by using VAE.
On the other hand, $\bm{m}_X, \bm{\sigma}_X$ in the latent distribution depend on the input image $X$, so static cues cannot be directly selected by eq.~\eqref{select_sc}.

\emph{iii}. 
By using NF, the encoded latent vector $Z$ is with expressive probability interpretation, which follows multivariate standard normal distribution $\mathcal{N}(\bm{0},I)$ independent of the input image $X$.
Thanks to the differentiable property of the NF model, the encoded latent vectors $Z_1, \cdots,  Z_L$ preserve the continuity over time.
Moreover, as experimentally shown in~(Dinh et al.~\citeyear{realnvp}; Kingma et al.~\citeyear{glow}), the latent space in NF encodes semantically meaningful concepts (like smile, blond hair, male, etc. on face dataset).
Because of these advantages, the flow-based approach instead of GAN and VAE is used to suppress static visual cues in our method.

\subsection{Integrated with Contrastive Learning}\label{suppress}
The proposed S$^2$VC method is  integrated in the framework of contrastive learning to obtain video representations less-biased to static cues.
In this work, positive pairs are constituted by the generated motion-preserved videos and corresponding inputs for self-supervised pre-training.
Given a video dataset $\mathcal{D}$ with $N$ samples $\mathcal{D} = \{V^1, V^2, ... , V^N\}$ for training, the loss function is defined as:
\begin{equation}\label{eqn:cl_loss}
\begin{footnotesize}
\begin{aligned}
    \mathcal L =-\mathbf{E}\left[\log\frac{\exp(v^{(i)}\cdot v_{p}^{(i)}/\tau)}{ \exp(v^{(i)}\cdot v_{p}^{(i)}/\tau) + \sum_{j \neq i}\exp(v^{(i)}\cdot v_{p}^{(j)}/\tau)}\right]
\end{aligned}
\end{footnotesize}
\end{equation}
where $\tau$ denotes the temperature parameter for model learning with hard negatives~\cite{instance_discrimination}.
In each positive pair, the motion-preserved video shares the same motion information as the original one but removes static cues.
By minimizing the loss function in eq.~\eqref{eqn:cl_loss}, the similarity of features in each positive pair is maximized.
Thus, the proposed method learns discriminative video representations which simultaneously preserve motion information and suppress static cues.
As an efficient and effective baseline, MoCo~\cite{moco} is employed for contrastive learning.
Furthermore, our method can serve as a powerful data augmentation technique and easily be integrated with other self-supervised learning methods, e.g., DPC in~(Han et al.~\citeyear{DPC}). 

%% file: sections/4-Experiments.tex
\section{Experiments}

\input{tables/4_1_1_sota}

\subsection{Datasets and Implementation Details}
Datasets used for experiments includes UCF101~(Soomro et al.~\citeyear{ucf101}), HMDB51~\cite{hmdb51}, Kinetics-400~\cite{k400}, and its subset Kinetics-200~\cite{s3d}.
We use the flow model as described in AdvFlow~(Dolatabadi et al.~\citeyear{advflow}) for video frame encoding and generation.
Two backbone networks, i.e., S3D~\cite{s3d} and R3D-18~(Hara et al.~\citeyear{r3d}), are evaluated for contrastive learning.
If not specified, we employ MoCo with S3D as the baseline and integrate our S$^2$VC with MoCo (optimized by eq.~\eqref{eqn:cl_loss} with $\tau$ set to 0.07).
For a fair comparison, we set the input clip length and resolution as 32, $128^2$ for S3D and 16, $112^2$ for R3D.
We conduct consistent augmentation for each frame in a video clip.
The batch size is set as 128 and the learning rate is initialized as 1e-3.
Total epochs we used for pretraining the network are 500 on UCF101, 200 on K200, and 100 on K400, respectively.
Please refer to the supplementary for more implementation details.

\subsection{Action Recognition}
We conduct self-supervised pre-training on two settings, i.e. linear probe and finetune.
For evaluation, following the common practice in~(Carreira et al.~\citeyear{I3D}; \citealt{wang2021multi}), we sample each video using half-overlap sliding window, and apply ten-crops test to each video clip. 
Then, we average the predicted accuracy as our validation result.
The results comparing with the state of the art are reported in Table~\ref{tab:sota_action_recognition_cmp}.

\noindent\textbf{Linear Probe.}
Follow the SimCLR~\cite{SimCLR}, we fix the weights of the pre-trained 3D CNNs and train a linear classifier after the last conv layer for 100 epochs.
We can observe from the last two columns in Table~\ref{tab:sota_action_recognition_cmp} that our method significantly surpasses existing works that use single RGB modality for pre-training.
Comparing with MemDPC~(Han et al.~\citeyear{MemDPC}), the improvement by our method is up to 11.9\% on UCF101 and 6.0\% on HMDB51.

\noindent\textbf{Finetune.}
We finetune the overall model for 500 epochs and show the results in Table \ref{tab:sota_action_recognition_cmp}. 
When the proposed S$^2$VC is introduced into MoCo, with the same backbone S3D and the same pre-train dataset UCF101, it can bring 5.2\% and 8.6\% improvements on UCF101 and HMDB51, respectively.
Due to limited computational resource, the S3D is pre-trained on K200 for only 200 epochs.
The results obtained under this setting have already been better than the CBT~\cite{sun2019learning} pre-trained on the larger scale K600+, and comparable with the SpeedNet~\cite{speednet} pre-trained on K400.
With the R3D pre-trained on UCF101, our method also achieves competitive performance and outperforms the VCOP~\cite{VCOP} pre-trained on the same dataset.
Though the CoCLR~(Han et al.~\citeyear{han2020self}) obtain higher accuracy than ours, it needs the additional optical flow modality complementary to RGB for pre-training.
Compared with the IMRNet~\cite{IMRNet} using multiple modalities in compressed videos for pre-training, our method achieves better results.
The performance gains by our method over the IMRNet are 4.4\% and 5.5\% respectively on the two datasets by using the same backbone and pre-training dataset K400.

\subsection{Video Retrieval}
In this section, our method is evaluated by the video retrieval task.
Following the setting in~\cite{VCOP}, we use the pre-trained 3D CNN with fixed weight as feature extractor.
The training set is defined as the \textit{gallery} and each 16-frame video clip from the test set is used as a \textit{query}.
If the category of the query appears in the retrieved $\mathcal{K}$-nearest neighbors, we record it as a \textit{hit} during the test time.
Accuracy comparison with other self-supervised learning methods on the UCF101 and HMDB51 is reported in Tables \ref{tab:recallatk_ucf101} and \ref{tab:recallatk_hmdb51}. 
When using the S3D as backbone, combining the S$^2$VC with MoCo brings a 4.1\% improvement on Top1 accuracy and 7.5\% improvement on Top5 accuracy on the UCF101 dataset.
For the HMDB51, the Top1 and Top5 gains are 1.7\% and 5.7\%, respectively.
Additionally, our method outperforms the state of the art for comparison, e.g., 6.2\% better than the BE~\cite{wang2020removing} under the same settings on the HMDB51.
These results validate that more discriminative and generalizable representations can be extracted by our method.

\input{tables/4_2_1_UCF101_Retrieval}
\input{tables/4_2_2_HMDB51_Retrieval}

\subsection{Ablation Study}
\input{figures/4_4_2_AlphaAblation}
\input{tables/4_3_1_Strategy}
\textbf{Motion Threshold $\alpha$.}
In our method, $\alpha$ in eq.~\eqref{select_sc} is an important hyperparameter to determine how many static cues are suppressed.
Retrieval results of different $\alpha$ are shown on the left of Fig.~\ref{fig:alpha_discussion}.
These results show that as $\alpha$ increases, the retrieval accuracy first increases, and then decreases after reaching the peak at 0.5 (the default value in this work).
Interestingly, when we select $\alpha$ as 0.8, which means only 6.5\%/5.7\% latent variables w.r.t motion are preserved in UCF101/HMDB51, 
the results are still better than small $\alpha$.
This indicates the importance of suppressing sufficient static cues and keeping conspicuous motion information for action recognition.
We also visualize the generation results of different $\alpha$ on the right of Fig.~\ref{fig:alpha_discussion}.
If $\alpha$ is too small, the effect for suppressing static cues is insufficient.
In contrast, for too large $\alpha$,  useful action cues may also be suppressed.

\noindent\textbf{Strategy for Suppressing Static Cues.}
In this experiment, we evaluate different Strategies for suppressing static cues.
First, we compare with a simple thresholding frame difference (TFD) method by pixel-level operation.
Similar to our method, each frame in a video is reshaped to $\mathbb{R}^{d}$ in the TFD.
Then, the top $20\%$ pixels with the largest STD along the time dimension are persevered (approximately equal to the amount of the preserved motion cues in the proposed S$^2$VC when $\alpha=0.5$).
Besides the TFD, three variants of the proposed S$^2$VC to determine the suppressed latent variables are evaluated:
(a) set to random noise: set latent variables in $C_s$ to normal noise for the first frame and keep them unchanged for other frames.
(b) shuffle - in clip: randomly shuffle each latent variable in $C_s$ between frames of a video.
(c) shuffle - in frame: randomly shuffle each latent variable in $C_s$ within a frame.
For fair comparison, we pre-train all the methods with the S3D on the UCF101 for 100 epochs.

Retrieval results are shown in Table \ref{tab:strategy}.
We have the following observations:
\emph{i}. The in-clip shuffle method brings little gain to the baseline model, since the channels already have similar values (small at standard deviation). 
\emph{ii}. All methods that strongly disturb the static cues show great improvement to the baseline model.
\emph{iii}. TFD surpass the MoCo baseline on UCF101 but perform worse on HMDB51.
This indicates that it is not robust and hard to generalize to a new dataset by simply detecting motion according to pixel level difference.
\emph{iv}. It performs the best by setting all latent variables in $C_s$ to zero.
Recall that zero suppressed latent variables refers to the minimum information entropy with the highest probability.
Other suppressing methods are not the most likely or with randomness to carry static information.
As a result, we use the S$^2$VC (set to 0) as default.

\subsection{Analysis on Suppressing Effects}
\noindent\textbf{Intra-class similarity of different samples.}
Given $\alpha=0.5$, we investigate the visual similarity over different samples in the same action category.
Specifically, we randomly select ten classes from UCF101/HMDB51 and sample a subset of video clips for each category.
Then, we measure the cosine similarity of different samples frame-by-frame with the latent vector $Z$ and the motion-preserved vector $Z_p$, respectively.
As frames in the same class may have a similar scene but large difference in moving regions, the cosine similarity is smaller if the latent vectors contain less static visual cues.
The decreased similarity of $Z_p$ compare with $Z$ shown in Fig.~\ref{fig:cos_sim} is aligned with the above analysis.
This phenomenon demonstrates that our method reduces the intra-class similarity of static object/background significantly, which ensures that the generated motion-preserved videos are less biased to static cues.
We also observe that different categories show widely varied ratios over $Z$ v.s. $Z_p$, which means action classes have various similarities on static cues.

\input{figures/4_4_3_CosSim}
\input{figures/4_4_4_Latent_Semantic}

\noindent\textbf{Visualization of the suppressing quality.}
More intuitively, we compare the generation of motion-preserved videos with minor/intense camera motion.
As shown in Fig.~\ref{fig:quality}, the generation is more robust to noise like camera movement and is able to focus on the most salient motion by suppressing static cues in the latent space encoded by the NF.

\subsection{Analysis on Performance Improvement}
\noindent\textbf{Relative Improvement over Static Classification.}
As our method suppress static visual cues, it may bring negative impact to classes that have a high correlation with non-moving object or background.
To study the correlation between the temporal dependency of actions and the performance gain brought by our method, we plot the class-level relative performance improvement in Fig.~\ref{fig:relative}.
In this experiment, we first train a randomly initialized S3D baseline using static videos (stacked copy images).
Since the stacked duplicate images provide no temporal information, the performance of the baseline model indicates how much a category depends on the static visual cues.
The plot shows that although there exist some classes that our method leads to a worse result, the overall performance is better.
Moreover, there is a clear negative relationship between relative gain and baseline performance, which suggests that the superiority of our method is mainly coming from precisely identifying actions with high temporal dependency.
We also find that our model shows a stronger negative correlation compared with MoCo.

\input{figures/5_1_RelativePerformance}
\input{figures/5_2_Heatmap}

\noindent\textbf{Salient Regions Compared with Optical Flow.}
In this experiment, we visualize the energy of the last convolutional layer with the Class-Activation Map (CAM) technique~\cite{zhou2016learning}. 
We sample from the HMDB51 instead of the UCF101 used for pre-training to show the generalizability.
We also visualize the optical flow for reference, which indicates the significant motion cues in the video.
The results are depicted in Fig.~\ref{fig:S$^2$VC_heatmap}.
From these samples, we find a strong correlation between highly activated regions and the dominant mover in the scene.
The network pre-trained with the S$^2$VC tends to focus more on the moving object. 
For example, in the second row, only our method concentrates on the two boys dribbling the ball on either side separately.

%% file: tables/4_1_1_sota.tex
\begin{table*}
    \centering
    {
    \begin{tabular}{p{0.24\linewidth}p{0.05\linewidth}P{0.05\linewidth}|p{0.06\linewidth}p{0.03\linewidth}P{0.05\linewidth}|p{0.065\linewidth}p{0.075\linewidth}|p{0.065\linewidth}p{0.075\linewidth}}
    \shline
    \multicolumn{3}{c}{\textbf{Method}}&\multicolumn{3}{c}{\textbf{Pretrain}}&\multicolumn{2}{c}{\textbf{Finetune}}&\multicolumn{2}{c}{\textbf{Linear Probe}}\\
    \hline
    {\bf Method}&{\bf Net}&{\bf Depth}&{\bf Dataset}&{\bf Res.}&{\bf +Mod.} &{\bf UCF101} & {\bf HMDB51} &{\bf UCF101} & {\bf HMDB51}\\
    
    \hline
    
    3D RotNet~\cite{jing2018self} & R3D & 17 & K400 & 112 & -  & 62.9 & 33.7 & 47.7 & 24.8 \\
    CBT~\cite{sun2019learning} & S3D & 23 & K600+ &112 & -  & 79.5 & 44.5 & 54.0 & 29.5 \\
    VCOP~\cite{VCOP} & R(2+1)D & 26 & UCF101 & 112 & -  & 72.4 & 30.9 & - & - \\
    DPC~(Han et al.~\citeyear{DPC}) & R2D3D & 33 & K400 & 128 & -  & 75.7 & 35.7 & - & - \\
    MemDPC~(Han et al.~\citeyear{MemDPC}) & R2D3D & 33 & K400 & 224 & -  & 78.1 & 41.2 & 54.1 & 30.5 \\
    SpeedNet~\cite{speednet} & S3D & 23 & K400 & 224 & -  & 81.1 & 48.8 & - & - \\
    RSPNet~\cite{RSPNet} & R3D & 17 & K400 & 224 & -  & 74.3 & 41.8 & - & - \\
    CoCLR~(Han et al.~\citeyear{han2020self}) & S3D & 23 & K400 & 128 & F  & 87.9 & 54.6 & 74.5 & 46.1 \\
    IMRNet~\cite{IMRNet} & R3D & 17 & K400 & 224 & M,R & 76.8 & 45.0 & - & - \\
    \hline
    {\bf MoCo Baseline} & S3D & 23 & UCF101 & 128 & -  & 69.3 & 35.1 & 46.6 & 21.4 \\
    {\bf Ours} & S3D & 23 & UCF101 & 128 & -  & 74.5\greenp{5.2$\uparrow$} & 43.7\greenp{8.6$\uparrow$} & 51.0\greenp{4.4$\uparrow$} & 27.7\greenp{6.3$\uparrow$} \\
    {\bf Ours} & S3D & 23 & K200 & 128 & -  & {\bf 82.5} & 48.4 & 63.8 & 35.9 \\
    {\bf Ours} & R3D & 17 & UCF101 & 112 & -  & 77.0 & 45.8 & 59.7 & 27.9 \\
    {\bf Ours} & R3D & 17 & K400 & 112 & -  & 81.2 & \textbf{50.5} & \textbf{66.0} & \textbf{36.5} \\
    
    \shline
    \end{tabular}
    }
    \caption{
    \textbf{Top-1 accuracy (\%) comparison with existing methods.}
    Action recognition results are reported on UCF101 and HMDB51 datasets.
    K200/K400/K600+ denote different versions of Kinetics.
    Res. is short for Resolution. 
    +Mod. means additional modalities besides RGB. 
    F is Optical Flow.
    M, R refer to the two modalities of P-frame in compressed videos.}
    \label{tab:sota_action_recognition_cmp}
\end{table*}

%% file: tables/4_2_1_UCF101_Retrieval.tex
\begin{table}[t]
\begin{tabular}{p{0.2\linewidth}P{0.08\linewidth}|ccccc}
\shline
\bf Method & \bf Net & \bf 1 & \bf 5 & \bf 10  & \bf 20 & \bf 50\\
\hline
SpeedNet & S3D & 13.0 & 28.1 & 37.5 &  49.5 & 65.0 \\
VCOP & R3D & 14.1 & 30.3 & 40.4 & 51.1 & 66.5 \\
MemDPC &R3D& 20.2 & 40.4 & 52.4 & 64.7 & - \\
Pace & R3D & 23.8 & 38.1 & 46.4 & 56.6 & 69.8 \\
\hline
MoCo & S3D & 32.8 & 49.0 & 57.5 & 68.3 & 80.7 \\
\textbf{Ours} & S3D & 36.9 & 56.5 & 65.6 & 75.0 & 86.3 \\
\textbf{Ours} & R3D & \textbf{39.9} & \textbf{57.1} & \textbf{66.3} & \textbf{75.6} & \textbf{87.4} \\
\shline
\end{tabular}
\caption{{\bf Recall-at-topK(\%)} of video retrieval on UCF101.}
\label{tab:recallatk_ucf101}
\end{table}

%% file: tables/4_2_2_HMDB51_Retrieval.tex
\begin{table}[t]
\begin{tabular}{p{0.2\linewidth}P{0.08\linewidth}|ccccc}
\shline
\bf Method & \bf Net & \bf 1 & \bf 5 & \bf 10  & \bf 20 & \bf 50\\
\hline
VCOP & R3D & 7.6 & 22.9 & 34.4 &48.8 & 68.9 \\
MemDPC &R3D&7.7&25.7&40.6&57.7&- \\
Pace & R3D&9.6 &26.9& 41.1& 56.1& 76.5\\
BE & R3D&11.9&31.3&44.5&60.5&81.4\\
\hline
MoCo & S3D & 13.2 & 31.8 & 44.0 & 59.7 & 80.7 \\
\textbf{Ours} & S3D & 14.9 & 37.5 & \textbf{51.7} & \textbf{68.3} & \textbf{84.5} \\
\textbf{Ours} & R3D & \textbf{18.1} & \textbf{37.9} & 51.1 & 66.0 & 84.4 \\
\shline
\end{tabular}
\caption{{\bf Recall-at-topK(\%)} of video retrieval on HMDB51.}
\label{tab:recallatk_hmdb51}
\end{table}

%% file: figures/4_4_2_AlphaAblation.tex
\begin{figure}[t]
    \centering
    	\includegraphics[width=\linewidth]{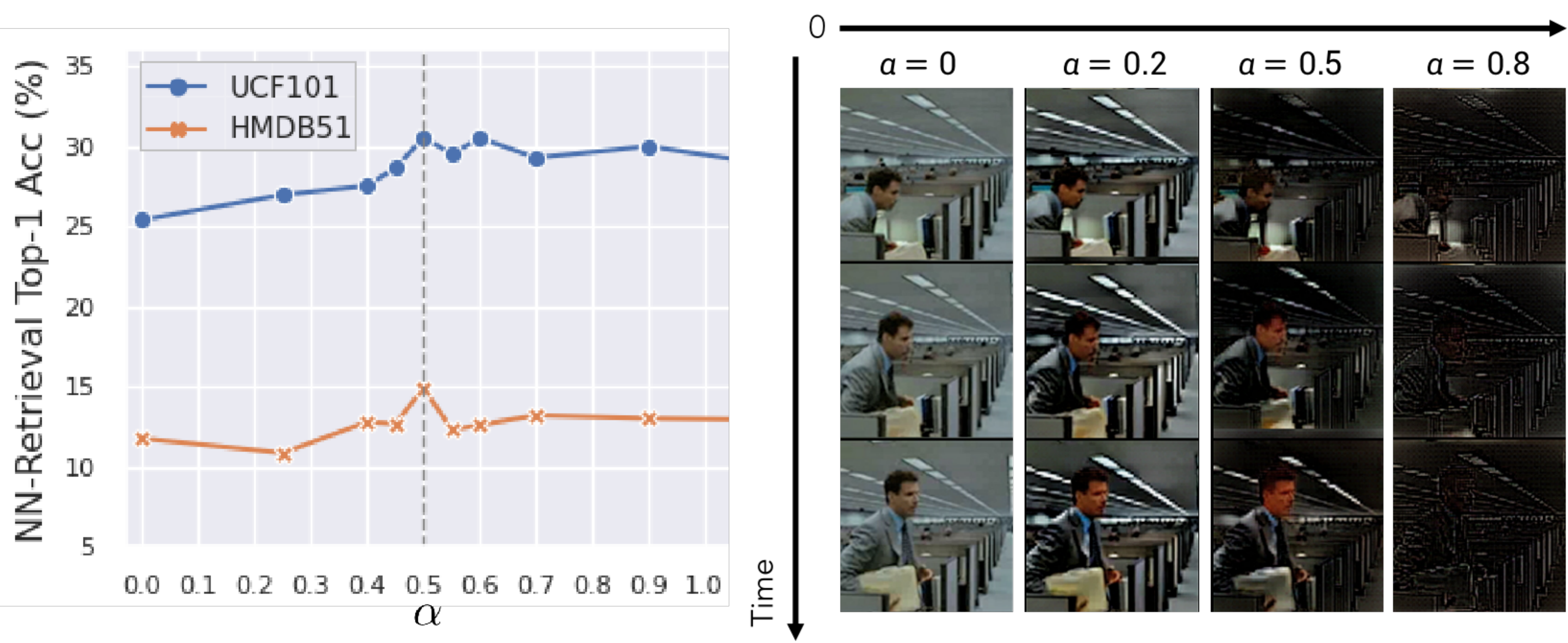}
\caption{
Left: R@1(\%) results of different $\alpha$ on UCF101 and HMDB51.
Best result achieved when $\alpha=0.5$ (only 22.7\%/22.3\% cues are preserved).
Right: Generation results of the same video with different $\alpha$.
For $\alpha$ larger than 0.8, almost all visual cues are suppressed. 
}
\label{fig:alpha_discussion}
\end{figure}

%% file: tables/4_3_1_Strategy.tex
\begin{table}[t]
\centering
{
\begin{tabular}{l|ll}
\shline
\multicolumn{1}{l|}{\textbf{Method}}&\multicolumn{1}{l}{\textbf{UCF101}}&\multicolumn{1}{l}{\textbf{HMDB51}}\\
\hline
MoCo & 25.4 & 11.8 \\
TFD & 29.1\greenp{3.7$\uparrow$} & 11.6\redp{0.2$\downarrow$} \\
S$^2$VC~(set to random noise)
 & 28.7\greenp{3.3$\uparrow$} & 12.6\greenp{0.8$\uparrow$}\\
S$^2$VC~(shuffle - in clip)
 & 25.8\greenp{0.4$\uparrow$} & 12.0\greenp{0.2$\uparrow$}\\
S$^2$VC~(shuffle - in frame)
 & 29.8\greenp{4.4$\uparrow$} & 14.0\greenp{2.2$\uparrow$} \\
\textbf{S$^2$VC~(set to 0) [default]}
 &\textbf{30.5}\greenp{5.1$\uparrow$} & \textbf{14.9}\greenp{3.1$\uparrow$} \\
\shline
\end{tabular}
}
\caption{R@1(\%) results of different strategies for suppressing static visual cues.
We use the S$^2$VC (set to 0) as default.}
\label{tab:strategy}
\end{table}

%% file: figures/4_4_3_CosSim.tex
\begin{figure}[t]
	\centering
	\includegraphics[width=\linewidth]{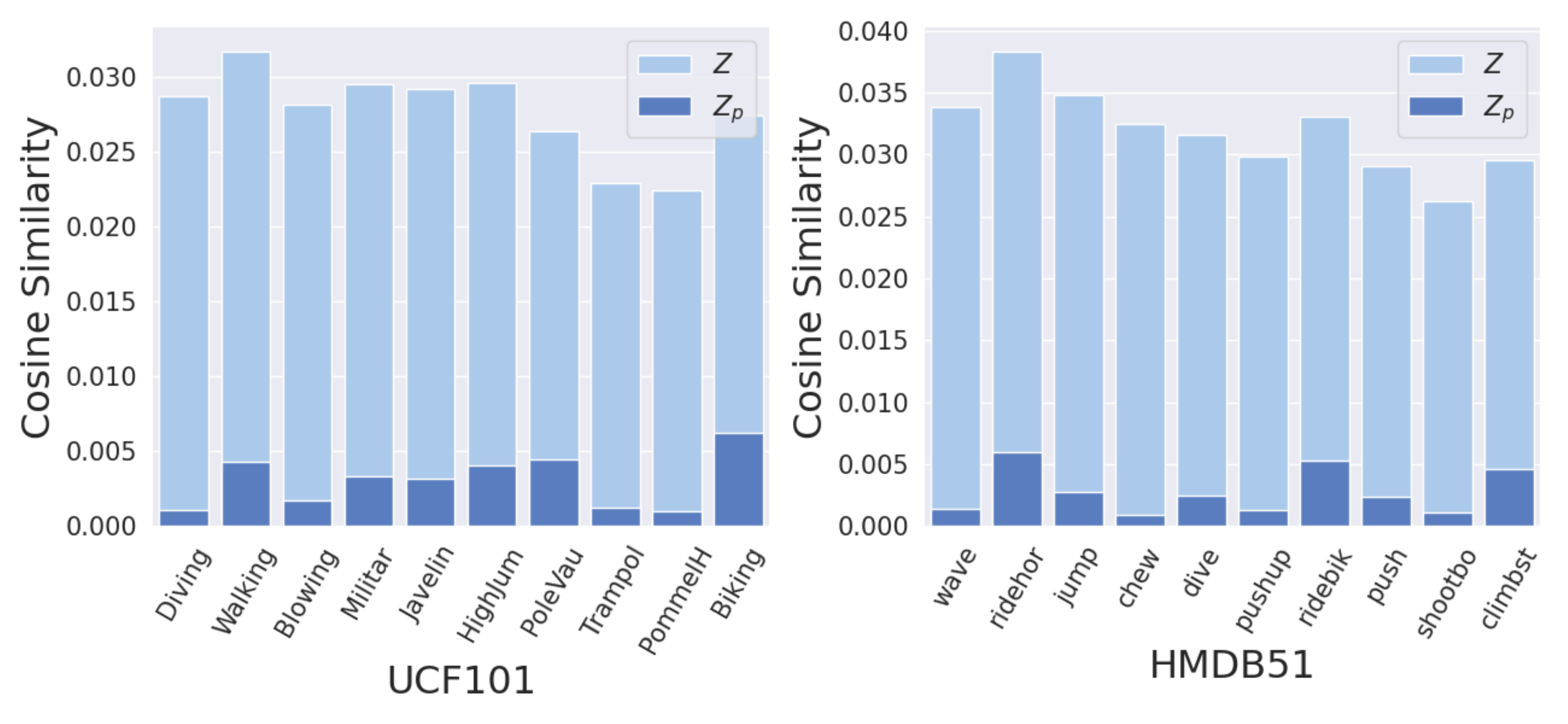}
	\caption{\textbf{Intra-class cosine similarity.} 
	Compared with $Z$, the similarity of the motion-preserved vector $Z_p$ greatly decreases, which indicates static cues are suppressed in $Z_p$.
	}
 	\label{fig:cos_sim}
\end{figure}

%% file: figures/4_4_4_Latent_Semantic.tex
\begin{figure}[t]
    \centering
    \includegraphics[width=.95\linewidth]{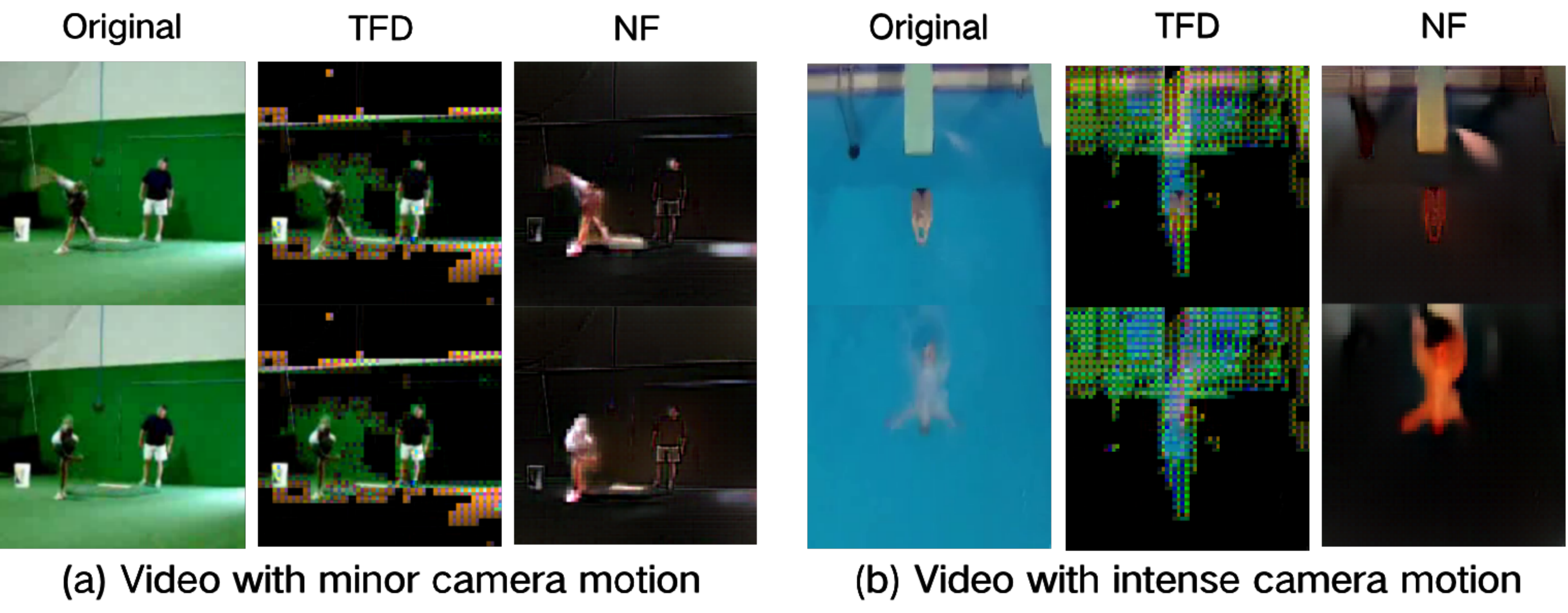}
\caption{
\textbf{Suppressing quality comparison} between the thresholding frame difference (TFD) with our method (NF).
}
\label{fig:quality}
\end{figure}

%% file: figures/5_1_RelativePerformance.tex
\begin{figure}[t]
	\centering
	\includegraphics[width=.75\linewidth]{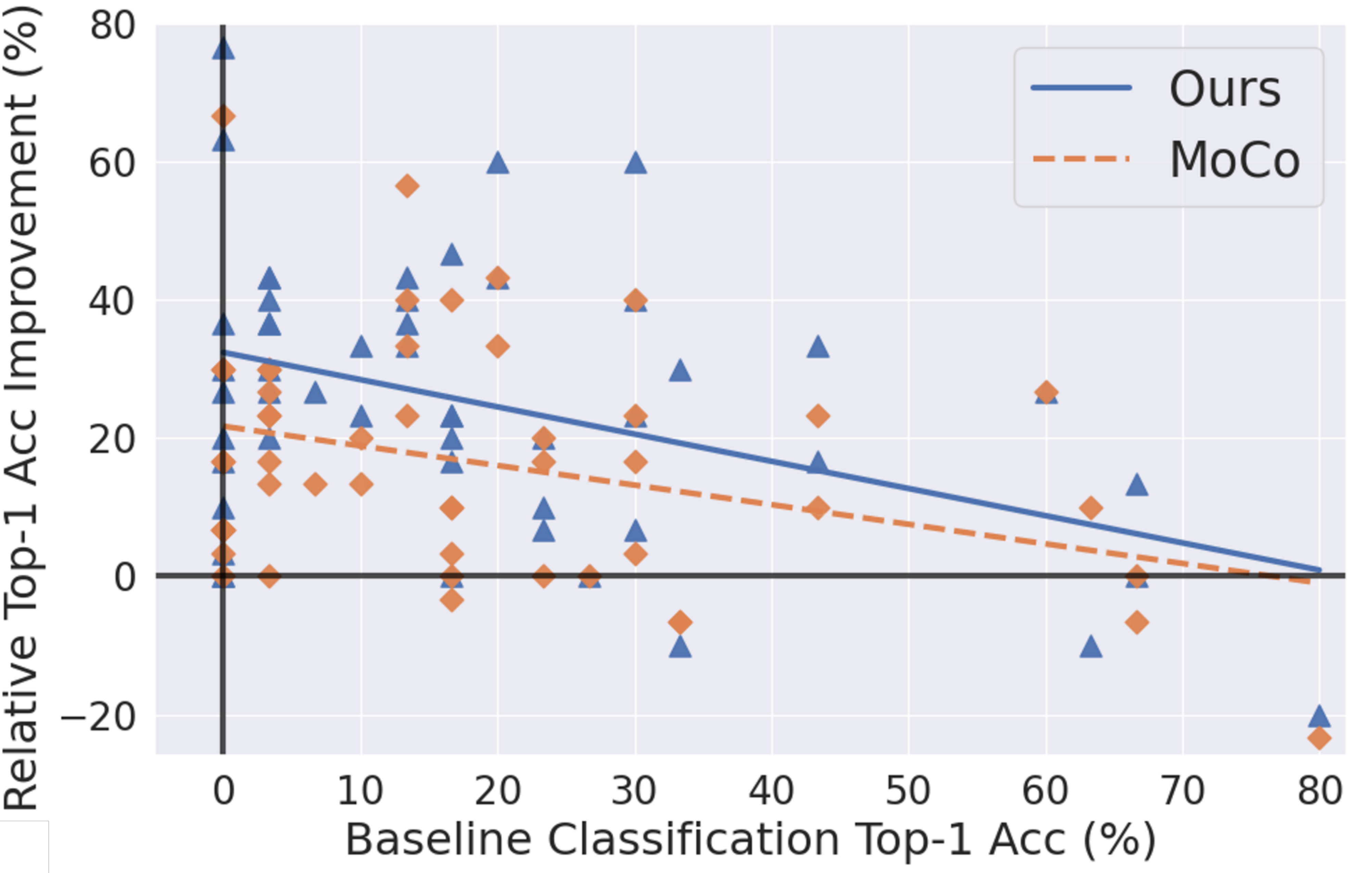}
	\caption{\textbf{Strong negative correlation between relative improvement and baseline classification result trained by static videos.}
	The Pearson correlation is $\rho = -0.40$ for \emph{S$^2$VC} and -0.33 for \emph{MoCo} baseline, which indicates S$^2$VC utilized more motion cues during discriminative learning.}
 	\label{fig:relative}
\end{figure}

%% file: figures/5_2_Heatmap.tex
\begin{figure}[t]
	\centering
	\includegraphics[width=.8\linewidth]{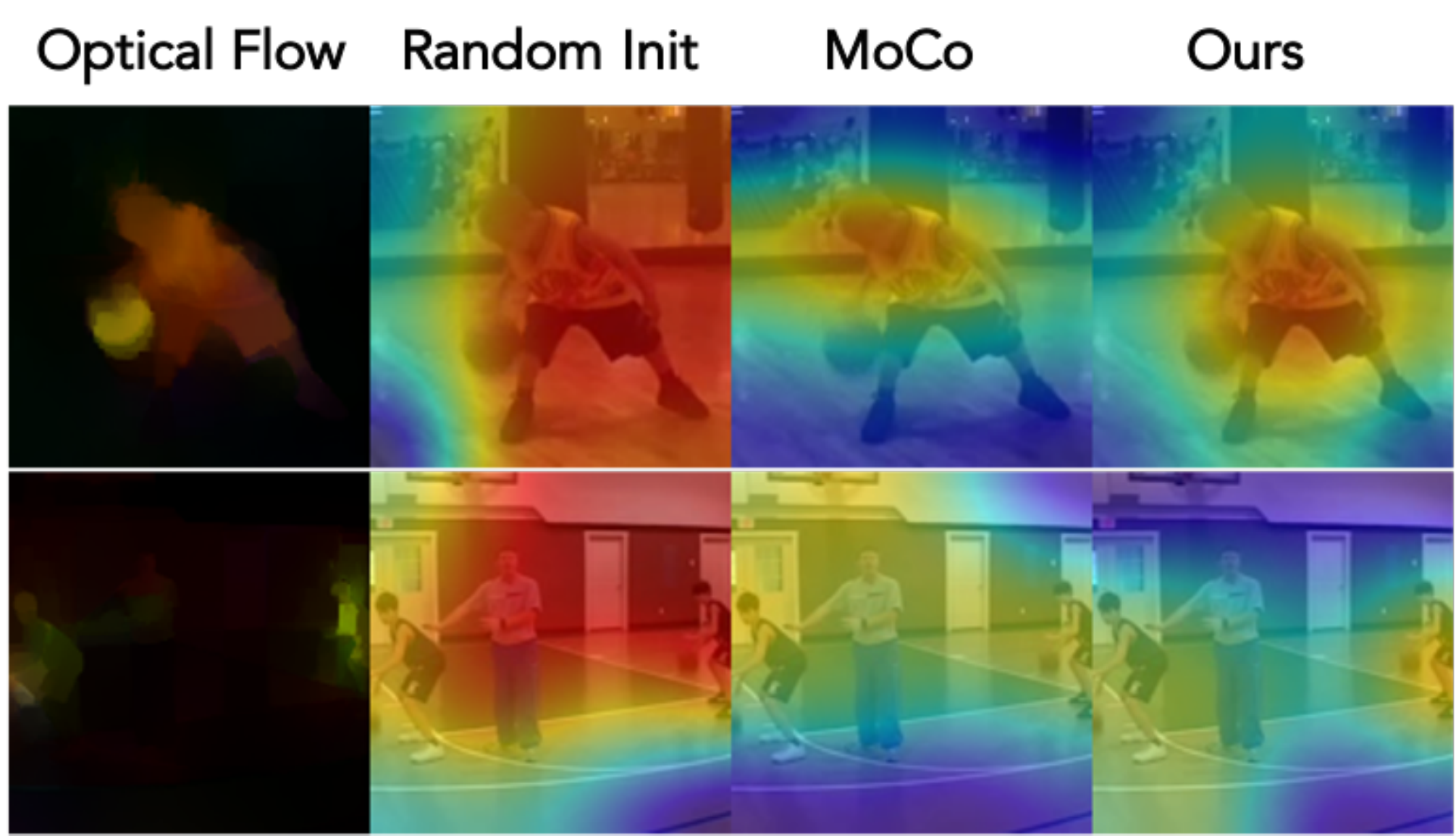}
	\caption{\textbf{Which region contributes most to identify action?} Here, red/blue correspond to high/low activated regions.
	Our method can discover the two salient moving players close to the boundary in the view (the second row).}
 	\label{fig:S$^2$VC_heatmap}
\end{figure}

%% file: sections/5-Conclusion.tex
\section{Conclusion}
In this paper, we present a novel method to suppress static visual cues (S$^2$VC), which mitigates the representation bias over less-moving object/background in videos.
Due to the difficulty in estimating the pixel-level distribution, video frames are encoded to a latent space under multivariate standard normal distribution by normalizing flows.
Then, less-varying latent variables along time are selected as static cues based on probabilistic analysis and suppressed to generate motion-preserved videos.
The proposed S$^2$VC is integrated with the self-supervised learning framework to extract video representations that focus more on \textit{motion cues}.
Extensive experiments with visualization validate that features learned  by our method pay more attention to moving objects and can be better generalized to different downstream tasks.

%% file: sections/S_3_pseudo_code.tex
\begin{algorithm}[h]
\BlankLine
\KwIn{A batch of $B$ video clips with $L$ frames, $C$ channels and spatial resolutions $H, W$, i.e., $D = \{V_1, \cdots, V_B \} \subset  \mathbb{R}^{L\times C\times H\times W}$.}
\BlankLine
\ForEach{\textnormal{video clip} $V \in D$}{
    \textcolor{gray}{\# down-sampling to reduced resolutions $H', W'$}\\
    $V'=adaptive\_avg\_pool2d(V, (H', W'))$\\
    \BlankLine
    \textcolor{gray}{\# suppressing static visual cues}\\
    \ForEach{\textnormal{video frame} $X \in V'$}{
        \textcolor{gray}{\# encoding to the $d$-dimensional latent space}\\
        $Z\ =f_\theta(X)\in\mathbb{R}^{d}, d=CH'W'$\\
    }
    \textcolor{gray}{\# selecting static cues $C_s$}
    $C_{s}=STD(Z_1,\dots,Z_L) < \alpha$\\
    \textcolor{gray}{\# generating the motion-preserved video $V'_p$} \\
    \ForEach{\textnormal{latent vector $Z$ w.r.t. each frame}}{
        $Z_p\leftarrow Z[C_{s}]=0$\\
        $X_p=f_\theta^{-1}(Z_p)$\\
        $V'_p \leftarrow X_p$ \\
    }
    \BlankLine
    \textcolor{gray}{\# capture information loss}\\
    $r = V - interpolate(V', (H, W))$\\
    \BlankLine
    \textcolor{gray}{\# up-sampling}\\
    $V_p = interpolate(V'_p, (H, W)) + r$\\
    }
\caption{Generating motion-preserved videos}
\label{algo:bg_suppress_detail}
\end{algorithm}
\linespread{1}

%% file: tables/S_5_DPC.tex
\begin{table}[h]
\centering
{
\begin{tabular}{l|ll|l}
\shline
\multirow{2}{*}{\textbf{Method}}&\multicolumn{2}{c}{\textbf{Recall-at-Top1(\%)}}&\multicolumn{1}{|c}{\textbf{Fine-tune}}\\

&\multicolumn{1}{l}{\textbf{UCF101}}&\multicolumn{1}{l}{\textbf{HMDB51}}&\multicolumn{1}{|l}{\textbf{UCF101}}\\
\hline
DPC & 7.1 & 3.5 & 45.7 \\
DPC+S$^2$VC & 9.3\greenp{2.2$\uparrow$} & 4.4\greenp{0.9$\uparrow$} & 47.5\greenp{1.8$\uparrow$}\\
\shline
\end{tabular}
}
\caption{Integrating S$^2$VC with DPC}
\label{tab:dpc}
\end{table}

%% file: figures/S_1_Statistics.tex
\begin{figure}[t]
	\centering
	\includegraphics[width=\linewidth]{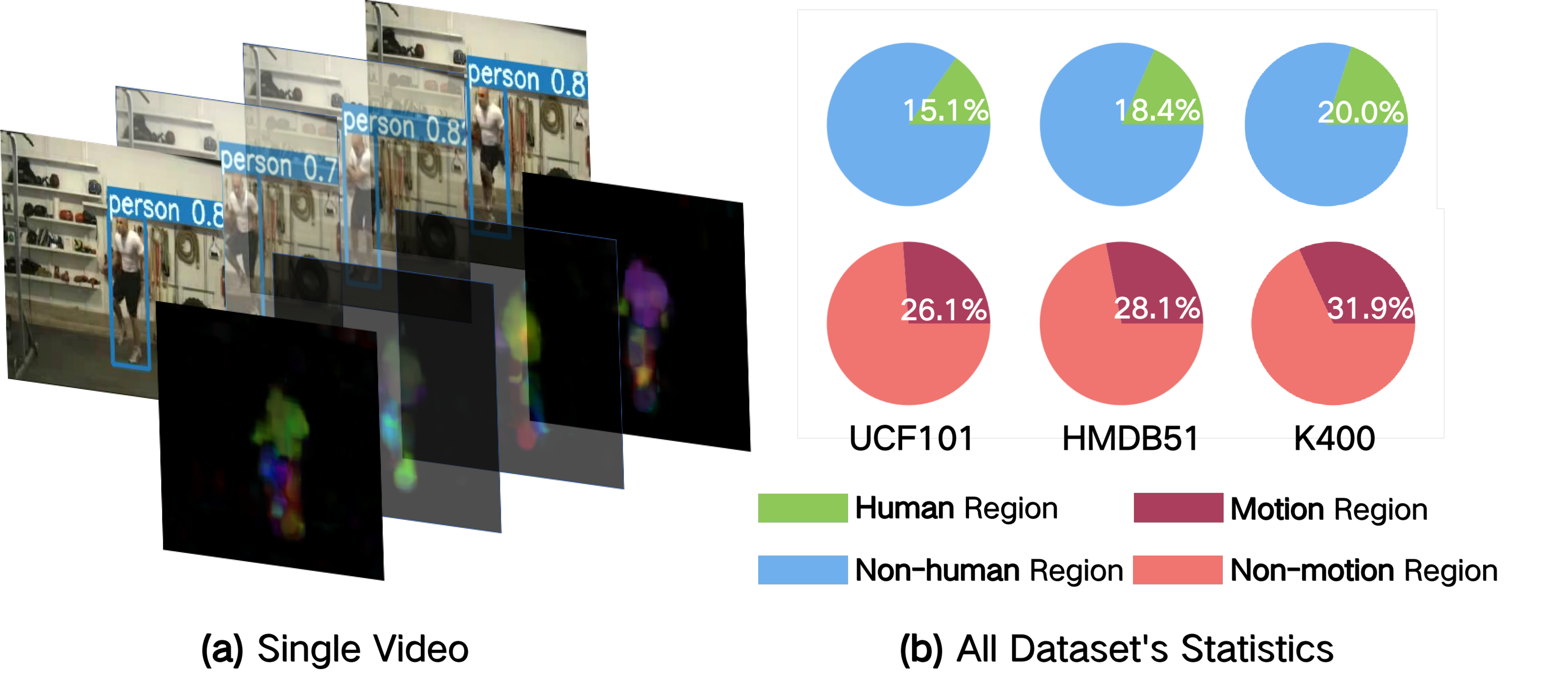}
	\caption{\textbf{The motion information in video is less dominated than static information}.
	It can be observed from (a) that both the human region and the motion region occupy only a small portion for the video, which is further confirmed by the statistic results shown in (b).
	}
 	\label{fig:staticstic}
\end{figure}

%% file: tables/S_2_Binary_Classification.tex
\begin{table}[h]
\centering
{
\begin{tabular}{ll|ll}
\shline
\multicolumn{1}{l}{\textbf{Method}}&\multicolumn{1}{l|}{\textbf{Pretrained}}&\multicolumn{1}{l}{\textbf{UCF101}}&\multicolumn{1}{l}{\textbf{HMDB51}}\\
\hline
Supervised & UCF101 & 68.1 & 73.5 \\
S$^2$VC & UCF101 & 73.8\greenp{5.7$\uparrow$} & 79.1\greenp{5.6$\uparrow$}\\
\hline
Supervised & K400 & 75.8 & 75.8 \\
S$^2$VC & K400 & 81.6\greenp{5.8$\uparrow$} & 81.0\greenp{5.2$\uparrow$}\\
\shline
\end{tabular}
}
\caption{Accuracy(\%) of identifying natural/shuffled videos on UCF101 and HMDB51 (Pre-trained 100 epochs).}
\label{tab:bin_test}
\end{table}

%% file: figures/S_6_Cos_Sim.tex
\begin{figure}[t]
	\centering
	\includegraphics[width=\linewidth]{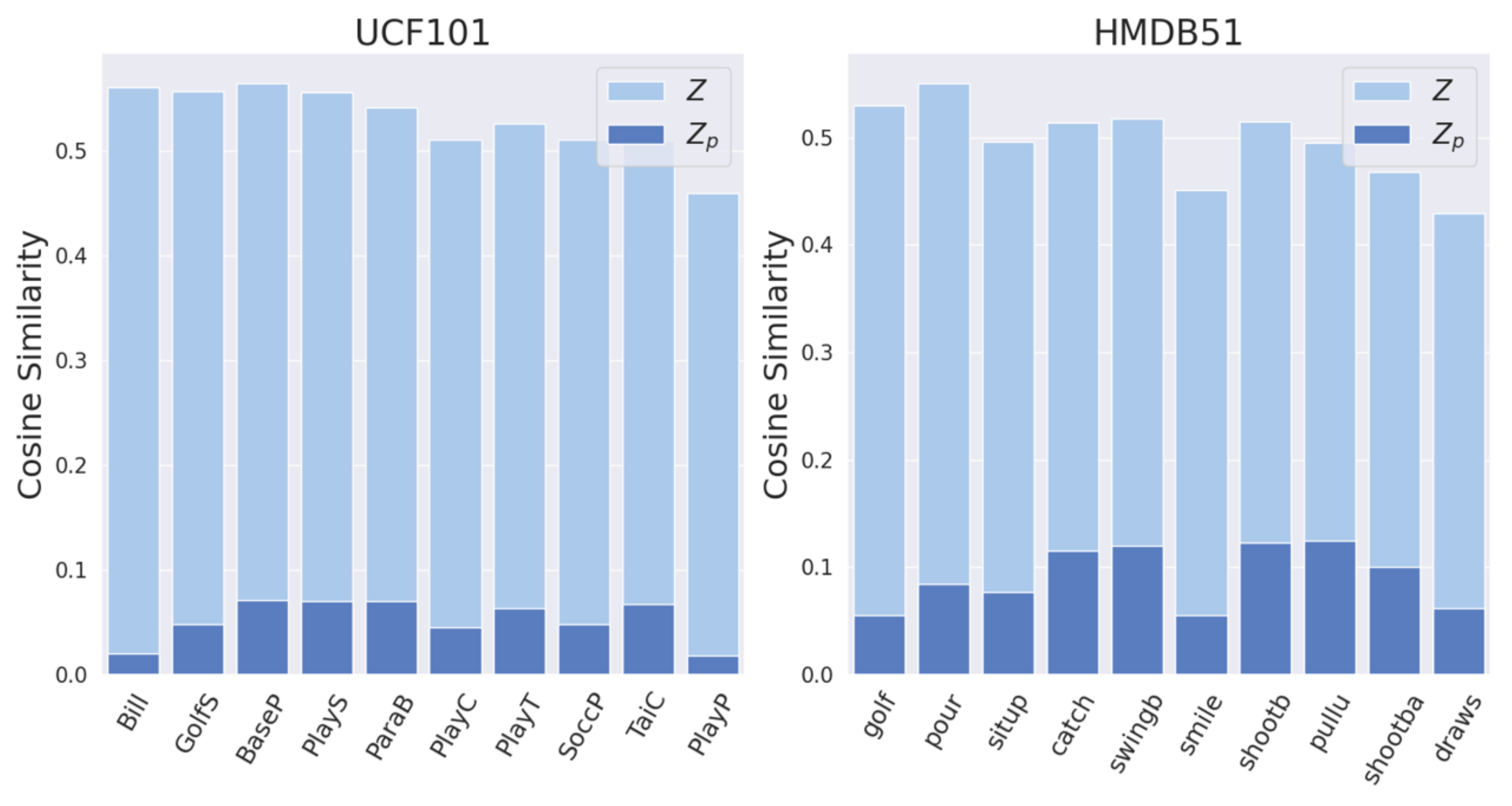}
	\caption{ Cosine similarity of intra-video comparison.}
 	\label{fig:intra_video_cos_sim}
\end{figure}
\begin{figure}[h]
	\centering
	\includegraphics[width=\linewidth]{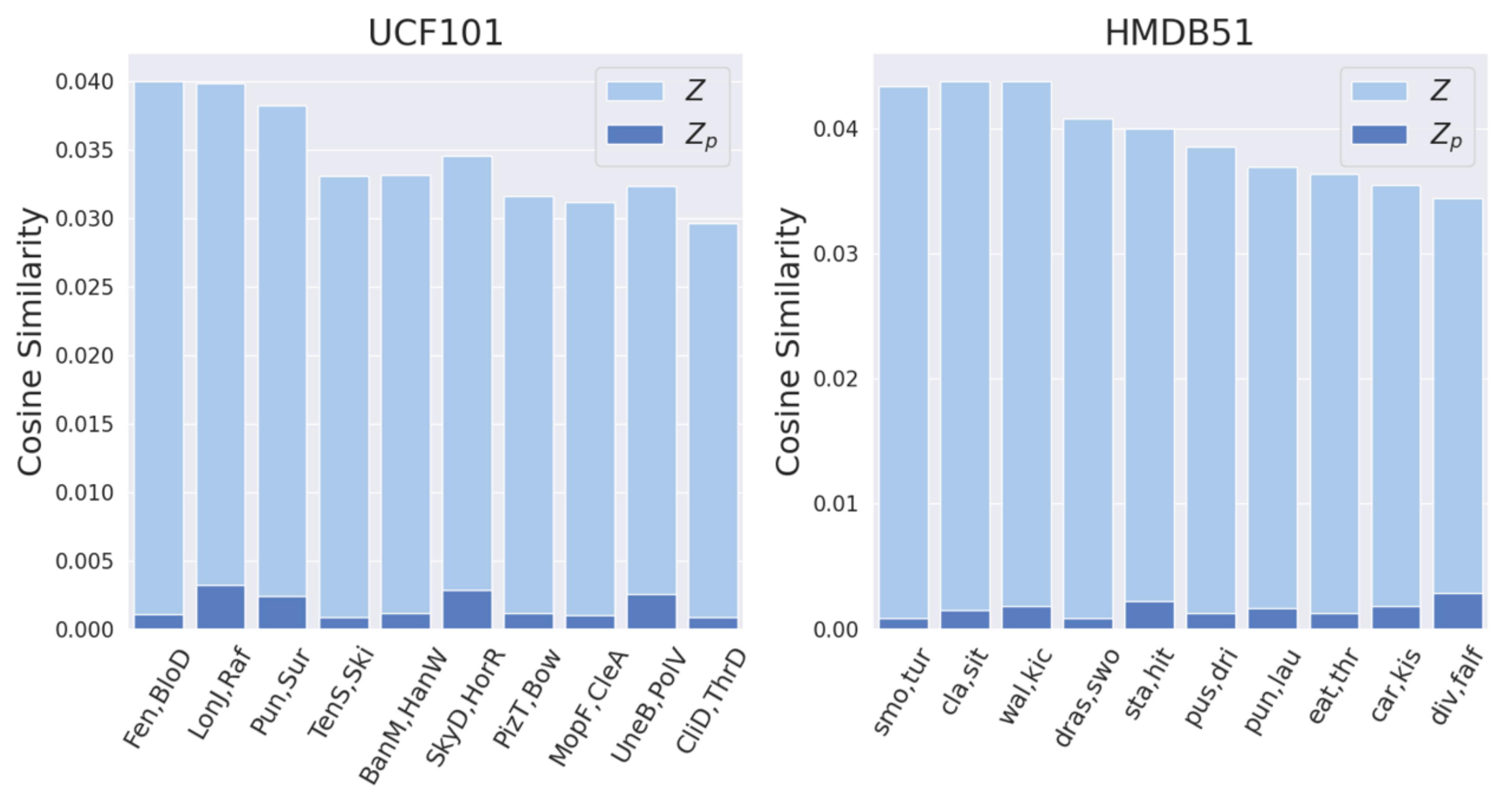}
	\caption{ Cosine similarity of inter-class comparison.}
 	\label{fig:inter_class_cos_sim}
\end{figure}

%% file: figures/S_7_Visualize_Sample.tex
\begin{figure}[t]
	\centering
	\includegraphics[width=\linewidth]{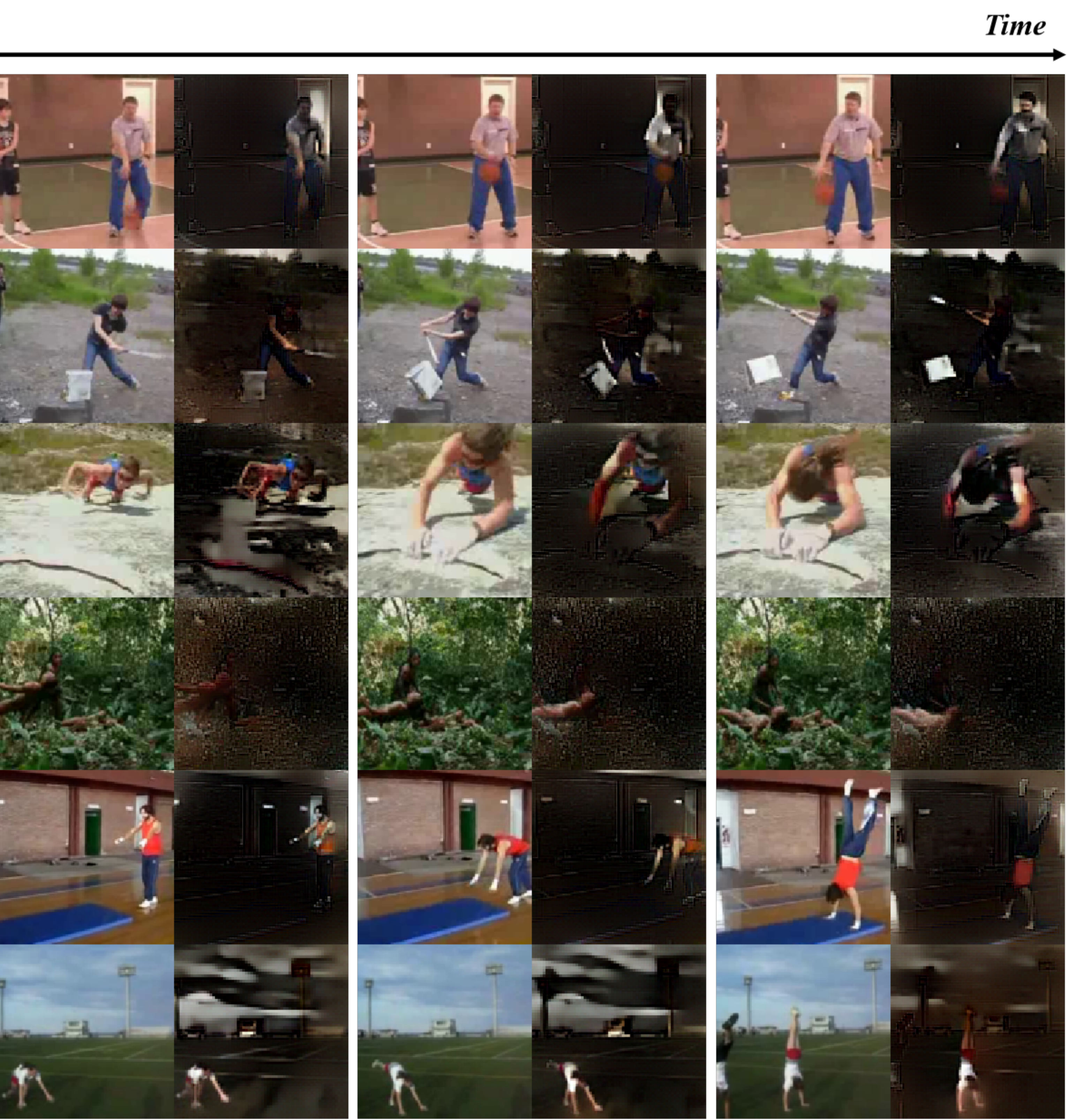}
	\caption{\textbf{Samples of suppressing results.}
	The left column is the original video and the right is the generated \textit{motion-preserved} video.
	The generated videos have the same motion as the original video, but most static cues are suppressed, such that the learned model can pay more attention to the salient mover.
	}
 	\label{fig:aug_visualize}
\end{figure}

%% file: figures/S_8_Norm_fit.tex
\begin{figure}[t]
	\centering
	\includegraphics[width=.85\linewidth]{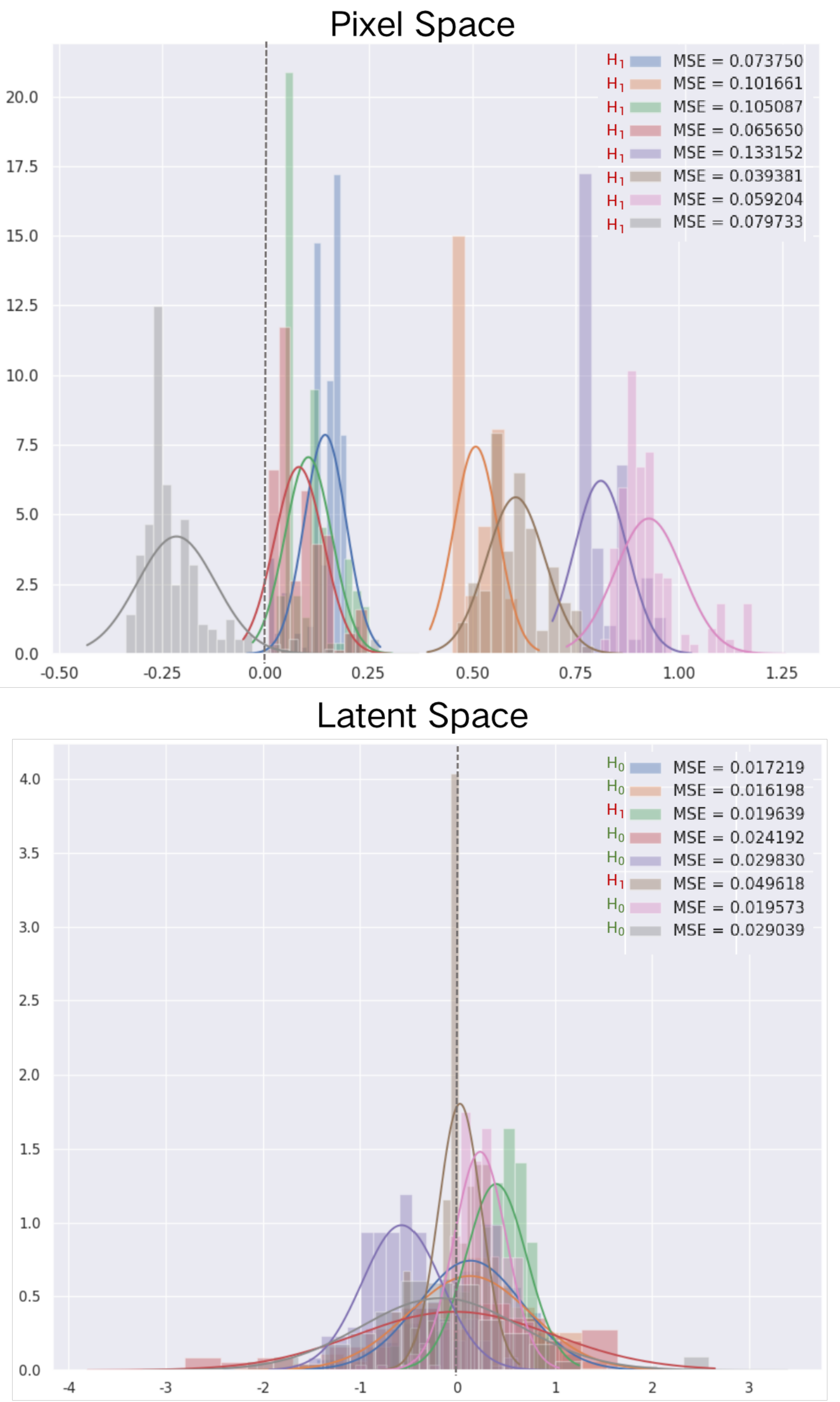}
	\caption{\textbf{Fitting normal distribution to pixels and latent variables conditioning on a video.}
	Latent variables conditioning on static factors in a video can be better fitting to normal distribution evaluated by the mean square errors (MSE)  and Kolmogorov-Smirnov test ($\alpha=0.05$).
	For the Kolmogorov-Smirnov test, $H_0, H_1$ denote the null hypothesis and rejection of null hypothesis, respectively.
	By using the proposed method, we cannot reject the null hypothesis for most latent variables which follow normal distribution.
	}
 	\label{fig:norm_fit}
\end{figure}